\input{styles/header.sty}

\title{HanjaBridge: Resolving Semantic Ambiguity in Korean LLMs via Hanja-Augmented Pre-Training}

\author{
  Seungho Choi\\
  Wisenut \\
  \texttt{csh1019@wisenut.co.kr}
}

\begin{document}

\maketitle


\begin{abstract}

Large language models (LLMs) often show poor performance in low-resource languages like Korean, partly due to unique linguistic challenges such as homophonous Sino-Korean words that are indistinguishable in Hangul script. To address this semantic ambiguity, we propose HanjaBridge, a novel meaning-injection technique integrated into a continual pre-training (CPT) framework. Instead of deterministically mapping a word to a single Hanja (Chinese character), HanjaBridge presents the model with all possible Hanja candidates for a given homograph, encouraging the model to learn contextual disambiguation. This process is paired with token-level knowledge distillation to prevent catastrophic forgetting. Experimental results show that HanjaBridge significantly improves Korean language understanding, achieving a 21\% relative improvement on the KoBALT benchmark. Notably, by reinforcing semantic alignment between Korean and Chinese through shared Hanja, we observe a strong positive cross-lingual transfer. Furthermore, these gains persist even when Hanja augmentation is omitted at inference time, ensuring practical efficiency with no additional run-time cost.
\end{abstract}


\section{Introduction}

Multilingual LLMs achieve impressive performance across various languages through joint training,~\cite{moosatransliteration} but these gains are not distributed equally among all languages.\cite{ogueji-etal-2021-small} In particular, languages with limited training corpora like Korean suffer a pronounced performance gap. During multilingual training, Korean often constitutes less than 0.1\% of the data in many open-source LLMs, leading to notably lower performance compared to high-resource languages. Dedicated Korean-centric models~\cite{kim-etal-2021-changes} have been developed to mitigate this gap, but they still do not fully resolve Korean’s unique linguistic challenges.


Approximately 57\% of Korean vocabulary is Sino-Korean (borrowed Chinese-origin words).\cite{park2019korean} These words share semantic and morphological roots with Chinese. For example, “연구” (Korean for “research”) shares the same characters as Chinese “\fixHanja{研}究” (yánjiū), and in both languages one can form perfectly equivalent sentences like “나는 오늘 \uline{연구}를 시작했다” (“I started my \uline{research} today”) and “我从今天开始\uline{\fixHanja{研}究}。” This phenomenon is possible because Korean and Chinese share a common logographic representation (Hanja), implying that cross-lingual semantic alignment can be relatively straightforward at the word level when using these characters. In contrast, English, which uses an alphabetic script, often requires context to interpret the correct meaning of a word that has multiple senses.


\begin{figure}
    \centering
    \includegraphics[width=1.0\linewidth]{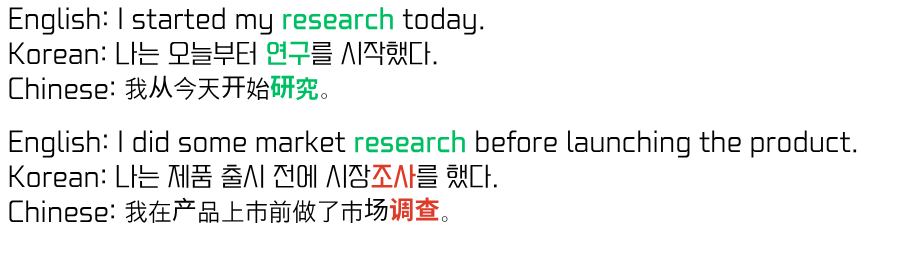}
    \caption{Context-dependent Sino-Korean translations of the English word research}
    \label{fig:research_alignment}
\end{figure}

\refig{fig:research_alignment}. Cross-lingual variations in the translation of the English word “research.” The same English word is translated as “연구(\fixHanja{研}究)” in a scholarly context, whereas it becomes “조사(\fixHanja{调查})” in a market-related context. This illustrates that although Chinese and Korean align semantically and morphologically via shared Sino-origin characters, English requires context-sensitive interpretation, making one-to-one mapping difficult. 

However, modern written Korean uses the phonetic Hangul alphabet, which does not distinguish between different Hanja that share the same pronunciation. Thus, many distinct Sino-Korean words have identical Hangul spellings, creating a high degree of polysemy and homophony in the language.\cite{park2019korean} In fact, about 35\% of Hanja entries in the Standard Korean Dictionary are homophones. For instance, the word “의사” can correspond to several different Hanja—“醫師” (doctor), “意思” (intention), “義士” (patriot), or “議事” (deliberation)—yet all are written “의사” in Hangul, making them virtually indistinguishable in text. This structural limitation of Hangul poses a major challenge for LLMs in accurately capturing word meanings and interpreting context. The model may struggle with semantic disambiguation since multiple meanings collapse into the same surface form.


To address these issues, we propose a Hanja-based semantic augmentation approach for Korean LLMs. The core idea is to leverage the shared logographic information from Chinese characters to inject explicit meaning cues into the model during training. By doing so, we aim to teach the model to resolve ambiguities that arise from Hangul’s one-to-many mapping between form and meaning. In the following, we detail our proposed method, HanjaBridge, which augments input text with Hanja characters and uses knowledge distillation to retain existing multilingual capabilities. We then demonstrate through experiments that this approach substantially improves the model’s understanding of Korean, especially on tasks requiring precise semantic comprehension, without sacrificing performance on English or other languages. We also analyze the model’s internal attention patterns to show that it learns to focus on context-appropriate meanings, validating the effectiveness of our approach.

\paragraph{Research Questions}

\textbf{RQ1}: Does the model leverage the provided Hanja characters in its internal attention, increasingly focusing on the correct Hanja candidates through training? In other words, as training progresses, does the model’s attention shift toward the true meaning among the listed Hanja forms? \textbf{RQ2}: When we combine token-level knowledge distillation with Hanja input augmentation, can we boost performance on Korean–Hanja alignment tasks while maintaining the model’s original inference efficiency (i.e. not degrading performance when Hanja are absent)? We will test whether this joint training improves Korean tasks without sacrificing English performance in inference, per the teacher’s guidance.





\section{Related Work}

\subsection{Continual Pre-Training for Low-Resource Languages}

Large multilingual pre-trained language models (PLMs) acquire broad knowledge across many languages from massive text corpora, but their performance on low-resource languages (LRLs) remains relatively poor\cite{liu-etal-2021-continual,guo-etal-2024-teaching}. To address this bias and strengthen a model’s expertise in a specific language, continual pre-training (CPT) has emerged as a key strategy\cite{fujinuma-etal-2022-match}. CPT involves further training an already pre-trained multilingual model on a large corpus of a target language, with the goal of improving its grasp of that language’s vocabulary, grammar, and contextual nuances. Recent works have successfully applied CPT to build high-performance models specialized for Korean~\cite{vo2024redwhale}, Japanese~\cite{fujiicontinual}, and Southeast Asian languages~\cite{dou-etal-2024-sailor}, indicating that CPT is essential for correcting model bias and enhancing representation for under-represented languages.


Early CPT approaches simply continued training on additional data in the target language. However, recent research has introduced more sophisticated strategies to improve efficiency and effectiveness. For example, to reduce the heavy computational cost of updating all model parameters, some works combine CPT with parameter-efficient fine-tuning (PEFT) techniques like LoRA.\cite{nag-etal-2025-efficient} There have also been proposals to optimize the training process, such as data importance reweighting~\cite{luo2024velocitune} and instruction-based CPT~\cite{chen2024instructioncp}, which dynamically adjust the curriculum or training focus. These developments suggest that CPT has evolved from simple data supplementation to designing an optimal training trajectory.


\subsection{Tokenizer Vocabulary Expansion}

A fundamental issue to address before effective CPT is the vocabulary mismatch problem in multilingual models. Most LLM tokenizers use subword algorithms (e.g., BPE), but their vocabularies are largely built from high-resource languages like English. Consequently, when processing text in languages with very different writing systems or morphology (such as Korean), a single meaningful morpheme or word often gets broken into multiple uninterpretable subword tokens – a phenomenon termed “semantic fragmentation.”~\cite{lee-etal-2024-length,kim-etal-2024-kombo} This fragmentation hampers the model’s ability to learn the cohesive meaning of words, inflating sequence length and diluting semantic signals.


To alleviate this, researchers have explored expanding or adapting the tokenizer’s vocabulary for the target language as part of CPT. By adding missing characters or frequent morphemes to the vocabulary, the model can represent important units as single tokens, preserving their meaning. An expanded vocabulary can reduce the number of tokens per word and thus reduce unnecessary complexity in learning word meaning.


\reftbl{tab:over-tokenization} shows a typical example of over-tokenization that occurs in Korean. When semantic units are segmented in this way during training, it becomes difficult for the model to maintain semantic consistency between words, and its contextual interpretation ability also deteriorates. Vocabulary expansion is the most direct way to structurally resolve this issue.


\begin{table}[h]
\centering\resizebox{\linewidth}{!}{
\renewcommand{\arraystretch}{1.6}
\begin{tabular}{|c|c|}
\hline
\makecell{\textbf{Original Word}} & \makecell{\textbf{Segmented Subwords}} \\
\hline
\makecell{\textcolor{red}{골칫}\textcolor{blue}{거리}\\(Nuisance / Headache)} & 
\makecell{\textcolor{red}{골칫} + \textcolor{blue}{거리}\\(Problem) + (Street / Distance)} \\
\hline
\makecell{\textcolor{red}{눈}\textcolor{blue}{치}\textcolor{orange}{채다}\\(Notice / Perceive)} & 
\makecell{\textcolor{red}{눈} + \textcolor{blue}{치} + \textcolor{orange}{채다}\\(Eye / Snow) + (Null) + (Snatch)} \\
\hline
\makecell{\textcolor{red}{입}\textcolor{blue}{덧}\\(morning sickness)} & 
\makecell{\textcolor{red}{입} + \textcolor{blue}{덧}\\(Mouth) + (Layer / Over)} \\
\hline
\end{tabular}
}
\caption{Examples of over-tokenization in Korean causing meaning distortion.}
\label{tab:over-tokenization}
\end{table}

\begin{figure*}[t]
    \centering
    \includegraphics[width=1\linewidth]{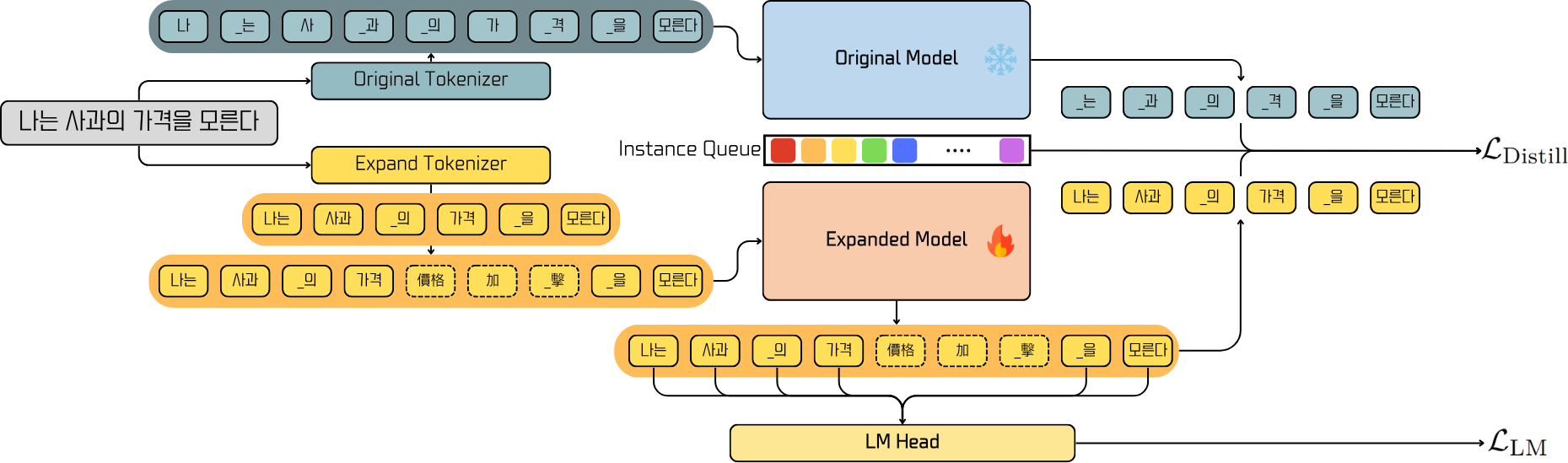}
    \caption{End-to-end workflow of our method.}
    \label{fig:main}
\end{figure*}

\subsection{Knowledge Distillation for Multilingual and Domain Adaptation}

Although CPT and vocabulary expansion markedly boost performance in the target language, they also risk \emph{catastrophic forgetting}: the model may overwrite parameters that encode general or cross-lingual knowledge, thereby degrading performance on other languages. This phenomenon represents the classic \emph{stability–plasticity dilemma} in continual learning. Unless it is mitigated, the gains obtained for one language can come at the cost of severe losses elsewhere, undermining the very purpose of a multilingual model.


To counteract catastrophic forgetting, a range of techniques have been explored, among which \textbf{knowledge distillation} (KD) has proven both effective and stable in the CPT setting.\cite{xu2024survey} In KD, the original, non-CPT model is treated as the \emph{teacher}, while the continually-trained model becomes the \emph{student}. The student is optimized to match the teacher’s output logits during CPT.\cite{hinton2015distilling} In this way, the student deepens its expertise in the new language while being explicitly regularized to preserve the teacher’s high-resource language knowledge.


The benefit of KD has been corroborated in recent LLM studies on low-resource language adaptation. For example, \cite{raju2025afroxlmr} showed that applying KD to African languages effectively suppresses catastrophic forgetting and preserves overall multilingual performance. Model-merging approaches have recently been proposed as alternative remedies~\cite{alexandrov-etal-2024-mitigating,tao-etal-2024-unlocking}, yet KD remains the most trusted and empirically validated technique when the training loop itself must enforce knowledge transfer. Our work therefore adopts KD within CPT to improve Korean performance while safeguarding the model’s capabilities in other languages.


\subsection{Resolving Semantic Ambiguity via External Signals}

A pipeline that combines CPT, vocabulary expansion, and KD forms a powerful technical framework for enhancing model performance in a particular language. Nonetheless, a purely training-centric approach cannot fully address the \emph{inherent semantic ambiguity} present in many languages. Korean, for instance, contains a large proportion of Sino-Korean words, and a single Hangul spelling can correspond to multiple Hanja forms—e.g., “의사” may mean \emph{doctor} (醫師), \emph{intention} (意思), or \emph{patriot} (義士). Without additional context, disambiguating such homophones is extremely challenging, making \emph{word-sense disambiguation} (WSD) a critical issue~\cite{10.1145/1459352.1459355}.


To tackle this problem, researchers have provided models with explicit \emph{external signals} or \emph{auxiliary information} that clarify word meaning. In the CJK family, machine-translation studies have shown that converting words to Hanja can dramatically improve translation quality—for Korean→Chinese~\cite{yoo-etal-2019-dont}, Korean→Japanese~\cite{kim-etal-2020-korean}, and Vietnamese→Chinese~\cite{li2020revisiting}. These works suggest that exposing deep etymological or semantic information, rather than surface forms alone, is a potent strategy for knowledge transfer.




\section{Method}
\label{sec:method}

To enhance the semantic expressiveness of a Korean LLM, we propose \textbf{HanjaBridge}, whose overall workflow is illustrated in \refig{fig:main}. Our approach consists of three key steps:  
(i) We constructed a joint Korean-Chinese Chinese character dictionary, then selected only Chinese characters corresponding to Korean prototypes and added them to a multilingual tokenizer.
(ii) These newly added Hanja tokens are \emph{appended in-line} to the sentence so that the model must resolve the correct meaning from context.  
(iii) We apply token-wise knowledge distillation, forcing the student model to mimic the teacher’s output distributions \emph{and} hidden representations, thereby boosting Korean performance without erasing existing multilingual knowledge.


\subsection{HanjaBridge: A Hanja-Augmented Semantic Injection Method}

\textbf{HanjaBridge} creates a single \emph{semantic slot} by concatenating \emph{all} candidate Hanja forms immediately after each ambiguous Hangul token.  
For example, the sentence “나는 사과의 가격을 모른다” (“I don’t know the price of the apple”) is internally expanded to  “나는 사과의 가격價格加擊을 모른다,” where 價格 (‘price’) and 加擊 (‘hit’) are plausible Hanja spellings for “가격.” During training, the model learns to select the contextually appropriate Hanja inside this slot, thereby disambiguating the original Korean word.


Because multiple Hanja candidates are presented simultaneously, the model cannot merely memorize a single correct answer; instead, it must leverage \emph{context} to decide which Hanja yields the highest-probability distribution.  
This not only trains the model’s innate disambiguation ability but also injects the rich semantic and etymological information encoded in high-resource Chinese directly into Korean tokens.  
Consequently, the model (1) interprets homophonous Korean words correctly in context, and (2) exploits Chinese semantic knowledge when processing Korean, leading to broad gains in expressiveness, reasoning accuracy, and knowledge coverage.


\begin{figure}[t]
    \centering
    \includegraphics[width=0.7\linewidth]{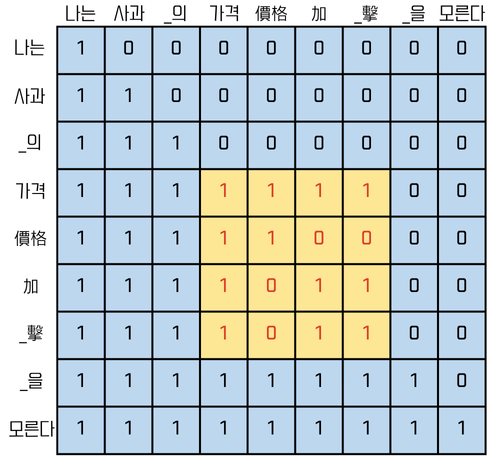}
    \caption{This is an illustration of the attention mask of HanjaBridge applied to the example sentence. Blue cells indicate the original baseline causal mask, while yellow cells highlight the additional connections introduced by HanjaBridge.}
    \label{fig:hanjabridge_local_attention}
\end{figure}

The model's attention mechanism is also adjusted to match these expanded inputs, as shown in \refig{fig:hanjabridge_local_attention}. 
Within each expansion group, the Korean token can freely attend to its Hanja candidates, but \emph{inter-candidate} attention is blocked.  
From a Korean perspective, these Hanja share the same surface form, yet from a Chinese perspective they differ semantically; allowing candidates to attend to one another would blur their meanings.  
By restricting the attention pathway to “Korean token ↔ each Hanja candidate,” we ensure that every candidate retains its independent meaning while still informing the Korean token’s representation.


During training, Hanja tokens participate \emph{only} as semantic hints; the language-modeling loss is computed exclusively on the original Korean tokens.  
Let the input length be \(L\) and assign a binary mask \(m_t \in \{0,1\}\) to each position \(t\).  
Positions with \(m_t=1\) are original tokens; \(m_t=0\) are Hanja expansions.  
Define the set of original positions as \(O=\{t \mid m_t=1\}\).  
Only hidden states \(h_t\:(t\in O)\) are fed into the LM head:


\begin{equation}
\mathbf{Z}_t=\mathbf{W}h_t+\mathbf{b},\qquad
P\!\bigl(y_t\mid\mathbf{x}_{<t}\bigr)=\operatorname{softmax}(\mathbf{Z}_t).
\label{eq:lm_head}
\end{equation}

\begin{equation}
\mathcal{L}_{\text{LM}}
      =-\sum_{t\in\mathcal{O}}
        \log P\!\bigl(y_t\mid\mathbf{x}_{<t}\bigr).
\label{eq:lm_loss}        
\end{equation}

Thus, Hanja tokens act only as latent cues via attention, while prediction and parameter updates focus on the original sequence.


\subsection{Token-wise Knowledge Distillation}

\begin{figure}
    \centering
    \includegraphics[width=1\linewidth]{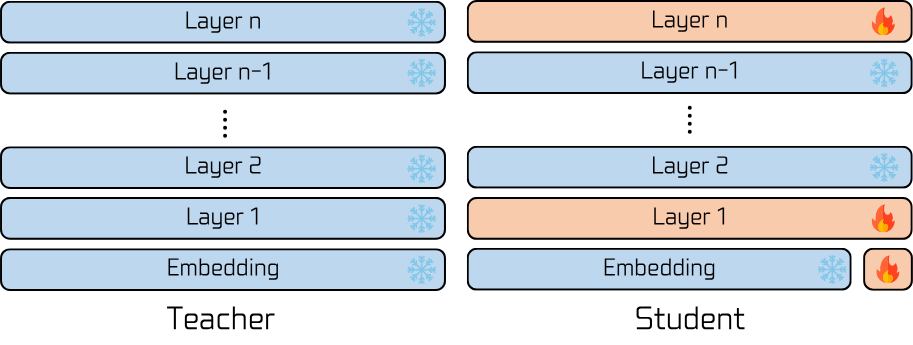}
    \caption{The training strategy for the Teacher-Student models. All parameters of the Teacher model are frozen and used solely as a knowledge source. The Student model is trained only on select layers and the newly expanded embeddings.}
    \label{fig:distillation_strategy}
\end{figure}

To preserve multilingual competence, we adopt token-level knowledge distillation (\refig{fig:distillation_strategy}).  
The pretrained source model serves as a \emph{frozen teacher}.  
The student is parameter-initialized from the teacher but unfreezes only selected layers—e.g.\ embeddings and a subset of Transformer blocks—enabling efficient training and mitigating catastrophic forgetting.


Within this teacher–student setup, we maintain an \emph{instance queue} that stores the teacher model’s token-level hidden vectors.  
The student is then trained to replicate these vectors at the feature level.  
Because the two models use different tokenizers, their sequence lengths can mismatch and the teacher may split a single word into multiple sub-words.  
We resolve this misalignment via \emph{offset mapping}: every original token position in the student’s input is put into one-to-one correspondence with the teacher, and when a teacher token is fragmented into several sub-words, we take the hidden vector of the \emph{last} sub-word as its representative.  
Hanja tokens serve only as auxiliary context and are therefore excluded from the distillation loss.


Our method distills knowledge by aligning \emph{token-level hidden representations}, not sequence-level summaries or output logits.  
This token-wise distillation offers two advantages.  
First, every token supplies a learning signal, enabling dense supervision.  
Second, even a single sentence adds dozens to hundreds of tokens to the queue, so the number of positive and negative examples grows explosively; this lets the student estimate the teacher’s feature space with far greater fidelity.  
As a result, the student learns to imitate not just individual vectors but the \emph{structure} of the feature space itself, preserving contextual meaning and relational information more effectively.


\begin{table*}[t]
\centering
\caption{
    Comparison of model performance on the main benchmarks.
    \textit{KoBALT} is a suite of difficult tasks that require deep linguistic understanding—syntax, semantics, and so on—whereas \textit{KoBEST} measures general natural-language-understanding (NLU) ability.  
    The variable $k$ denotes the maximum number of Hanja candidates supplied during HanjaBridge training.  
    Boldface marks the highest average (\textit{avg}) score within each benchmark group.
}
\label{tab:main_results}
\resizebox{\textwidth}{!}{%
\begin{tabular}{@{}l|cccccc|cccccc|ccccc@{}}
\toprule
\multirow{2}{*}{\textbf{Model}} & \multicolumn{6}{c|}{\textbf{Korean (KoBALT - Hard)}} & \multicolumn{6}{c|}{\textbf{Korean (KoBEST - General)}} & \multicolumn{5}{c}{\textbf{English}} \\
\cmidrule(lr){2-7} \cmidrule(lr){8-13} \cmidrule(lr){14-18}
& syntax & semantics & pragmatics & morphology & phonetics & avg & boolq & copa & hellaswag & sentineg & wic & avg & boolq & copa & hellaswag & wic & avg \\
\midrule
Qwen2.5-3B & 0.1067 & 0.2047 & 0.0988 & 0.0714 & 0.1129 & 0.1189 & 0.5413 & 0.6340 & 0.4600 & 0.4962 & 0.4229 & 0.5108 & 0.7722 & 0.8500 & 0.5505 & 0.6176 & 0.6976 \\
Full CPT & 0.1000 & 0.1907 & 0.0741 & 0.0476 & 0.1452 & 0.1115 & 0.7236 & 0.6450 & 0.4460 & 0.6080 & 0.5887 & 0.6023 & 0.7471 & 0.8200 & 0.5020 & 0.5313 & 0.6501 \\
\midrule
k=0 (distill only) & 0.0900 & 0.1581 & 0.0741 & 0.0952 & 0.0968 & 0.1028 & 0.7201 & 0.6470 & 0.4560 & 0.5668 & 0.4550 & 0.5690 & 0.7731 & 0.8300 & 0.5089 & 0.5893 & 0.6753 \\
k=2 & 0.1033 & 0.2000 & 0.0998 & 0.0238 & 0.0645 & 0.0983 & 0.7137 & 0.6520 & 0.4480 & 0.5743 & 0.6085 & 0.5993 & 0.7865 & 0.8200 & 0.5068 & 0.5846 & 0.6745 \\
k=4 & 0.1033 & 0.2186 & 0.1358 & 0.0238 & 0.1129 & 0.1189 & 0.6830 & 0.6600 & 0.4500 & 0.6650 & 0.6064 & 0.6129 & 0.7662 & 0.8200 & 0.5087 & 0.5172 & 0.6531 \\
\textbf{k=8 (Ours)} & 0.1067 & 0.2279 & 0.1358 & 0.1190 & 0.1290 & \textbf{0.1437} & 0.7244 & 0.6560 & 0.4500 & 0.7380 & 0.6087 & \textbf{0.6354} & \textbf{0.7788} & \textbf{0.8300} & \textbf{0.5060} & \textbf{0.6019} & \textbf{0.6792} \\
k=16 & 0.1067 & 0.2140 & 0.1235 & 0.0714 & 0.1452 & 0.1322 & 0.6980 & 0.6700 & 0.4480 & 0.6952 & 0.6086 & 0.6240 & 0.7725 & 0.8300 & 0.5059 & 0.5846 & 0.6733 \\
\bottomrule
\end{tabular}%
}
\end{table*}

To steer the student’s feature learning effectively, we adopt a \emph{contrastive, queue-based distillation} strategy.\cite{fang2021seed}
We maintain a fixed-length instance queue $D$: after each mini-batch, the teacher model’s token-level hidden vectors are enqueued, and the oldest entries are dequeued so that the queue size stays constant.  
During training, for every original token $i$ we align the teacher vector $z^{T}_{i}$ with the corresponding student vector $z^{S}_{i}$.  
We then form the augmented set $D^{+}=D\cup\{z^{T}_{i}\}$—that is, the current teacher vector temporarily pushed onto the queue—and compute the similarity distributions $p_{T}(i)$ and $p_{S}(i)$ as defined in \refeq{eq:sim}.  
When $z_{i}=z^{T}_{i}$ we obtain $p_{T}$; when $z_{i}=z^{S}_{i}$ we obtain $p_{S}$.  
The student consults \emph{only} teacher vectors, never its own past outputs, and minimizes the cross-entropy between $p_{T}(i)$ and $p_{S}(i)$.  
This procedure supplies a rich \emph{multi-point alignment} signal: the student learns not merely to copy each teacher vector in isolation, but also to reproduce its \emph{relative relationships} to every other token in the queue, enabling more comprehensive knowledge transfer.


\begin{equation}
p(i,j) \;=\;
\frac{\exp\!\bigl((z_i \cdot d_j)/\tau\bigr)}
     {\sum_{d \in D^+} \exp\!\bigl((z_i \cdot d)/\tau\bigr)}.
\label{eq:sim}
\end{equation}

The overall training procedure jointly optimizes two losses:  
(1) the standard language-modeling loss for the student, $L_{\text{LM}}$, and  
(2) the token-level distillation loss $L_{\text{KD}}$ defined in \refeq{eq:kd_loss}.  
Both losses are computed \emph{only} at positions corresponding to the original Korean tokens; Hanja tokens are excluded from training.


\begin{equation}
L_{\text{KD}} = -\frac{1}{N} \sum_{i=1}^{N} \sum_{d_j \in D^+} p_T(i,j) \, \log p_S(i,j).
\label{eq:kd_loss}
\end{equation}
 
The final objective is the weighted sum of the two losses, as shown in \refeq{eq:total_loss}:


\begin{equation}
L_{\text{total}} = L_{\text{LM}} + \lambda L_{\text{KD}}.
\label{eq:total_loss}
\end{equation}

Here, $\lambda$ is a hyper-parameter that balances the gradient magnitudes of the two terms.  
This parallel optimization enables the student to internalize not only raw language-modeling ability but also the teacher’s representational structure.  
Empirically, the proposed distillation scheme yields larger performance gains than simple logit matching, demonstrating that the combination of Hanja-based auxiliary information with dense feature supervision is an effective vehicle for knowledge transfer.


\section{Experiments}

\subsection{Experimental Setup}

\paragraph{Hardware.} All training was conducted on a system with 4× NVIDIA H200 80GB GPUs, using mixed precision (BF16) training for efficiency.

\paragraph{Base model.} We build on the open-source \textbf{Qwen 2.5-3B} model as our base. We apply our proposed \texttt{HanjaBridge} augmentation, tokenizer expansion, and token-level distillation to this model as described in \refsec{sec:method}.

\paragraph{Continual pre-training data.} We use only \textbf{Korean data} for continual pre-training. (No additional English or Chinese data was used in this phase.) The corpus is drawn from the FineWeb-Edu dataset~\cite{penedo2024fineweb}, a high-quality Korean text collection.

\paragraph{Hyper Parameters.} We set the sequence length to $65{,}536$ tokens. For knowledge distillation, we use a temperature of $\tau_T = 0.01$ for the teacher and $\tau_S = 0.2$ for the student, and a distillation loss weight of $\lambda = 0.1$.

\paragraph{Benchmarks.}
\begin{itemize}
\item \textbf{Korean:} KoBEST~\cite{jang2022kobest}, KoBALT~\cite{shin2025kobalt}
\item \textbf{English:} BoolQ~\cite{clark-etal-2019-boolq}, COPA~\cite{roemmele2011choice}, Hellaswag~\cite{zellers-etal-2019-hellaswag}, WiC~\cite{pilehvar-camacho-collados-2019-wic}
\end{itemize}

KoBEST is an evaluation suite of 5 Korean downstream tasks requiring linguistic knowledge, constructed with high-quality human-generated data. KoBALT is a comprehensive Korean language understanding benchmark covering five linguistic domains (syntax, semantics, pragmatics, phonology, morphology). For English, we use widely adopted benchmarks that test commonsense reasoning and contextual understanding: BoolQ, COPA, HellaSwag, and WiC. All tasks are evaluated in zero-shot mode (no fine-tuning), and accuracy is used as the evaluation metric for all tasks.


\subsection{Main Results: Korean vs.\ English}

We evaluate our proposed HanjaBridge method on both Korean and English benchmarks to assess overall performance. We compare the optimal configuration (our model with $k=8$ Hanja candidates) against the original pre-trained model and two baselines: a standard continual pre-training without our semantic augmentation (Full CPT), and a CPT with knowledge distillation but no Hanja augmentation (k=0). \reftbl{tab:main_results} summarizes the results.


\paragraph{Korean Language Understanding}: HanjaBridge substantially improves the model’s Korean understanding abilities, especially on tasks requiring nuanced semantic disambiguation. On the challenging KoBALT-Hard benchmark that evaluates deep linguistic knowledge, our \texttt{k=8} model achieves an average score of 0.1437, outperforming the baseline (0.1189) by about 21\% relative improvement. This demonstrates that providing Hanja candidates helps the model grasp complex syntactic and semantic structures more effectively. Similarly, on the more general natural language understanding suite KoBEST-General, our model records the highest average score of 0.6354, outperforming all other models. These results validate our hypothesis that explicitly injecting semantic candidates via HanjaBridge is a highly effective strategy for enhancing Korean-specialized capabilities.


\paragraph{Preventing Catastrophic Forgetting and Cross-Language Transfer} One of the key issues in continuous learning for language specialization is catastrophic forgetting, where models lose their ability to understand other languages. Without separate mitigation strategies, the \texttt{Full CPT} model shows a significant decline in English performance from an average of 0.6976 to 0.6501, clearly demonstrating the problem of forgetting. In contrast, our methodology successfully addresses this issue. The \texttt{k=8} model not only demonstrates outstanding performance in Korean but also effectively preserves English performance, achieving an average score of \textbf{0.6792}. This score is significantly higher than the benchmark of the \texttt{Full CPT} model and even slightly exceeds the performance of the \texttt{k=0 (distill only)} model. This demonstrates that meaning enhancement through Hanja does not come at the expense of English proficiency.


\paragraph{Effect of the number of Chinese character candidates (\texttt{k})} Through experiments, we also discovered a trend regarding the number of Chinese character candidates (\texttt{k}). The Korean benchmark performance improved overall as \texttt{k} increased from 2 to 8. However, this trend did not continue indefinitely, and the \texttt{k=16} model showed a slight performance decline compared to the \texttt{k=8} model. This suggests that while providing candidate options is beneficial, an excessive number of candidates can introduce noise and complicate the model's semantic discrimination process.


\begin{figure}
    \centering
    \includegraphics[width=0.8\linewidth]{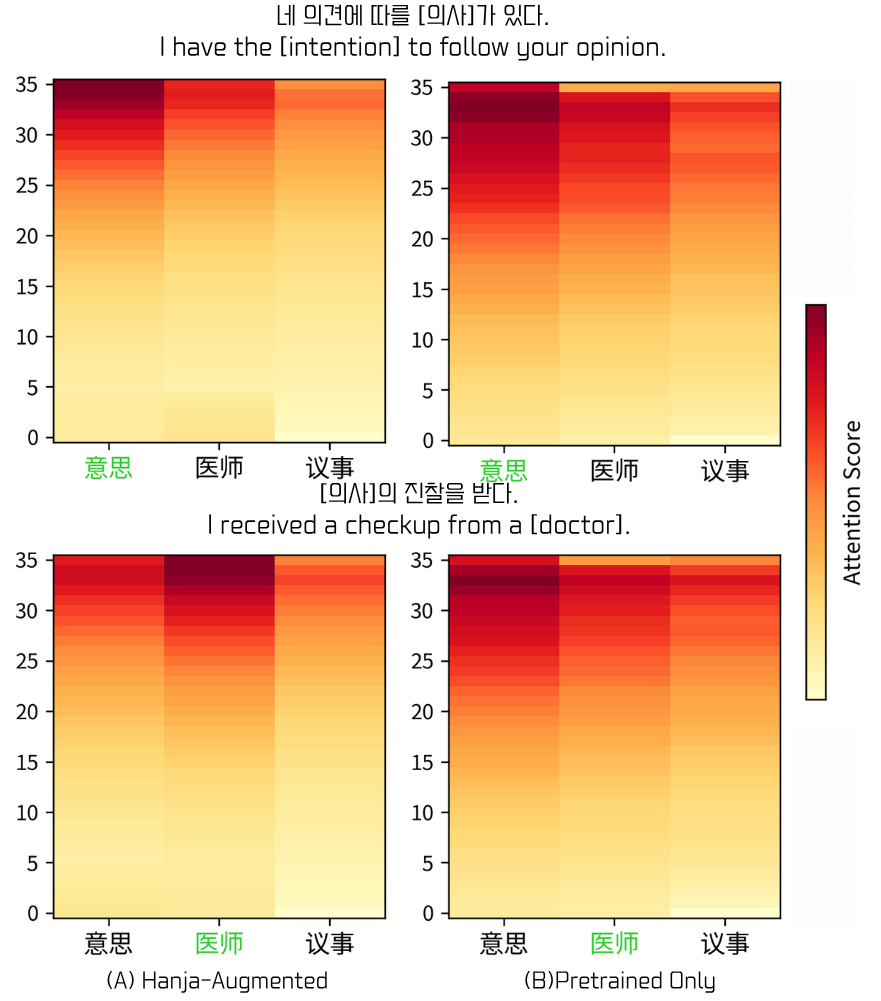}
    \caption{Cumulative attention heat-maps (36 heads × 12 layers) for two sentences containing the ambiguous Korean word “의사.” Green highlights mark the context-appropriate Hanja (醫師 ‘doctor’; 意思 ‘intention’). (A) HanjaBridge focuses strongly on the correct character, whereas the (B) Pretrained-Only model distributes attention almost uniformly across all candidates.}
    \label{fig:attn_heatmap}
\end{figure}

\subsection{RQ1 – Attention-focus Analysis}

In RQ1, we analyzed whether the proposed Chinese character pairing technique was correctly reflected in the model's internal attention. To verify this, we analyzed the self-attention weight patterns of the model trained after Chinese character pairing. Specifically, we observed whether the model's attention was focused on the correct candidate among the multiple Chinese character candidates input together with the Korean word. For example, in the sentence “I don't know the price of apples,” we examined whether the Hanja token 價格 (price) or 加擊 (hit) within the 價格 expansion group was assigned a higher attention weight based on the context.

However, the raw attention map of the Transformer layer tends to be flattened, making direct interpretation difficult. This is because, as the layers increase, each token's embedding becomes dependent on the overall context, making it difficult to understand how information is propagated using simple weights alone. Therefore, instead of simple attention probabilities, we utilized propagation-based contribution analysis methods such as Attention Rollout. This method calculates the cumulative effect of attention across multiple layers from input to output, ultimately tracking the extent to which a specific input token contributed to the formation of the target output embedding. 

\refig{fig:attn_heatmap} shows the results of visualizing the Attention Rollout of the HanjaBridge model and the original (Pretrained-Only) model after accumulating 36 layers, using two Korean sentences containing homophonic Chinese characters such as “의사 (physician/intention/deliberation).” All Hanja candidates following the Korean tokens were input, and the green characters indicate the correct Hanja for each context. The model trained using Hanja annotation focuses attention more strongly on the correct Hanja characters (醫師, 意思) in both sentences, while the original model distributes attention almost evenly across all candidates. This visually supports the claim that expanded Hanja tokens actually enhance meaning representation.




\begin{table}[t]
\centering
\resizebox{\linewidth}{!}{%
\begin{tabular}{c|cccccc}
\toprule
\multirow{2}{*}{$k$ (Number of Candidates)} & \multicolumn{6}{c}{Training Steps} \\
\cmidrule{2-7}
& 30k & 60k & 90k & 120k & 150k & 180k \\
\midrule
2 & 0.573 & 0.581 & 0.597 & 0.611 & 0.631 & 0.638 \\
3 & 0.391 & 0.413 & 0.427 & 0.446 & 0.453 & 0.477 \\
4 & 0.321 & 0.338 & 0.347 & 0.362 & 0.375 & 0.404 \\
5 & 0.267 & 0.286 & 0.324 & 0.349 & 0.360 & 0.388 \\
6 & 0.108 & 0.146 & 0.209 & 0.267 & 0.323 & 0.361 \\
7 & 0.055 & 0.108 & 0.125 & 0.186 & 0.234 & 0.285 \\
8 & 0.102 & 0.148 & 0.177 & 0.239 & 0.277 & 0.308 \\
\bottomrule
\end{tabular}%
}
\caption{RQ1: Accuracy of selecting the correct Hanja candidate across training steps. $k$ indicates the number of Hanja candidates, and values represent the proportion where the highest attention was assigned to the correct candidate.}
\label{tab:rq1_accuracy}
\end{table}

\reftbl{tab:rq1_accuracy} shows the results of quantitative evaluations performed on tens of thousands of sentences using experiments such as \refig{fig:attn_heatmap}, demonstrating how much attention the model focuses on the correct character candidates (correct character tokens) during the training steps (30k $\sim$ 180k). The values in the table represent the “accuracy” (the percentage of correct candidates receiving the highest attention) within each difficulty level ($k$ = number of candidates) interval.

The experimental results show that accuracy steadily increases in all cases as learning progresses. For example, in the easy interval ($k=2$), accuracy increased from 0.573 to 0.638 (+6.5 percentage points), and in the difficult interval ($k=8$), accuracy increased from 0.102 to 0.308 (+20.6 percentage points), confirming that the model gradually utilizes Hanja expansion information more actively. Additionally, differences in convergence speed depending on difficulty are observed. In the easy case ($k=2$), the improvement in performance slows down after a certain amount of learning, while in the difficult case, the upward trend continues even at the end of learning. This suggests that convergence should be considered for each difficulty level when designing the learning schedule.

In summary, the RQ1 experiment strongly supports the hypothesis that the proposed method actually utilizes Hanja information at the attention level and that the tendency to focus on correct Hanja characters is reinforced as learning progresses. In particular, the consistent improvement in high-difficulty samples supports the core claim that Hanja-Bridge and token-level distillation effectively enhance the semantic discrimination ability of Korean LLMs.




\subsection{RQ2 – Hanja Multiple-choice Probe}

\paragraph{Evaluation Setup} We measured whether the model could select the correct Chinese character notation corresponding to \emph{homophonic Korean words} based on context. The prompt format is as shown below, and the model is tasked with solving a multiple-choice question by selecting the correct Chinese character from $k$ options (candidate Chinese characters) in a zero-shot manner.

The model score is calculated using the log-likelihood \(log_p\) for each option. To eliminate \textbf{bias due to option order}, $k$ prompts were generated so that the correct candidate appeared once at each position, and the average log-likelihood for each option was used as the final score.



\begin{figure}
    \centering
    \includegraphics[width=1\linewidth]{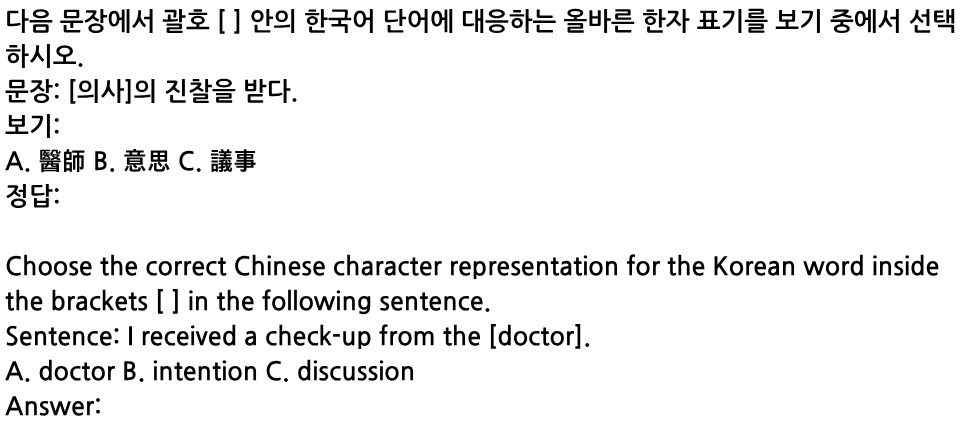}
    \caption{Examples of evaluation prompts ($k=3$).}
    \label{fig:rq2_prompt}
\end{figure}

\paragraph{Comparison Settings}
\begin{itemize}
\item \textbf{HanjaBridge (w/ HB-inf.)}  : HanjaBridge applied to both learning and inference
\item \textbf{HanjaBridge (without HB-inf.)} : HanjaBridge applied during training, inference uses Korean only
\item \textbf{Full CPT} : Full-parameter CPT performed using Korean corpus only
\item \textbf{Pretrain} : Original pre-trained model Qwen2.5-3B
\end{itemize}


\paragraph{Analysis and Implications}
(i) In \reftbl{tab:rq2_main}, the proposed model outperforms pretraining (0.503) by 6.5 percentage points and Full CPT (0.389) by 17.9 percentage points even when only using Chinese character combinations in the training stage (0.568). In other words, once Hanja merging is used to inject semantic discrimination information during training, it maintains its advantage during inference without requiring additional tokens. This means that both \textbf{inference efficiency}—no increase in token length—and \textbf{model versatility}—can be deployed as-is on existing APIs and servers—are secured simultaneously.

(ii) Pre-trained models already possess a certain level of Korean–Chinese character meaning connection, but \emph{Chinese character merging CPT} further \textbf{concretizes and reinforces this at the lexical level}, raising it to the highest value of 0.575. This provides quantitative evidence that the tendency observed in RQ1, where attention focuses on the correct Chinese character, also translates into actual task performance.

(iii) The setting that uses merged tokens for both learning and inference is slightly higher (+0.7 percentage points), but the difference is not significant, so in actual service, the “merge only during learning” mode is a reasonable choice in terms of cost-performance balance.

In summary, RQ2 demonstrates that Hanja merging CPT is a practical method that enhances Korean semantic resolution while allowing the existing tokenizer and system to be used as-is during the inference stage.





\begin{table}
\centering
\resizebox{\linewidth}{!}{%
\begin{tabular}{|l|c|c|c|c|}
\hline
 & HanjaBridge (w HB-inf.)
 & HanjaBridge (w/o HB-inf.) 
 & Full CPT
 & Pretrain \\
\hline
acc & 0.575 & 0.568 & 0.389 & 0.503 \\
\hline
total token & 961,982 & 922,384 & 922,384 & 1,030,151 \\
\hline
avg token per sample & 96.19 & 92.23 & 92.23 & 103.01 \\
\hline
\end{tabular}%
}
\caption{Accuracy comparison based on Chinese character merging method and inference conditions}
\label{tab:rq2_main}
\end{table}







\section{Conclusion}

In this study, we proposed a new continuous pre-training framework called \textbf{HanjaBridge} to solve the semantic ambiguity problem that large language models (LLMs) encounter due to the unique linguistic characteristics of Korean, particularly homophones in Sino-Korean words. HanjaBridge does not inject a single definitive Hanja character but instead provides all possible Hanja candidates in parallel, guiding the model to actively discern meaning through context. This process is accompanied by token-level knowledge distillation and vocabulary expansion to prevent catastrophic forgetting.

The experimental results clearly demonstrate the high effectiveness of HanjaBridge. The proposed technique achieved remarkable performance improvements on Korean benchmarks, particularly achieving a \textbf{21\% relative performance improvement} on the KOBALT benchmark, which requires deep language understanding. At the same time, knowledge distillation successfully preserved existing multilingual capabilities, especially English performance, maximizing cross-language transfer effects. Through attention analysis, we quantitatively confirmed that the model gradually focuses on the correct Hanja candidates that match the context during the learning process.

One of the most important practical contributions of this study is that such performance improvements are achieved without additional computational costs during inference. Even if Hanja is only used during training and omitted during inference, performance is maintained, ensuring high efficiency that can be seamlessly integrated into existing systems.

In conclusion, HanjaBridge goes beyond simple data augmentation and introduces a new paradigm that leverages the inherent ambiguity of language as a learning signal. This approach not only raises the performance of Korean LLMs to a new level but also has the potential to be widely applied to improve the performance of other low-resource languages with similar notation systems.

HanjaBridge directly depends on the quality and coverage of the constructed Hanja-Hangul mapping dictionary. Therefore, there is an inherent limitation in that the ability to distinguish meaning may be limited for neologisms or specialized terms not listed in the dictionary. To overcome these limitations and expand the research, it is necessary to systematically analyze the impact of the overall scale and quality of the Hanja dictionary on model performance. This will enable the establishment of guidelines for constructing dictionaries optimized for specific domains and the exploration of the optimal balance between cost and performance.
 
\clearpage
\bibliography{bibliography}

\begin{thebibliography}{32}
\providecommand{\natexlab}[1]{#1}

\bibitem[{Alexandrov et~al.(2024)Alexandrov, Raychev, M{\"u}ller, Zhang, Vechev, and Toutanova}]{alexandrov-etal-2024-mitigating}
Alexandrov, A.; Raychev, V.; M{\"u}ller, M.~N.; Zhang, C.; Vechev, M.; and Toutanova, K. 2024.
\newblock Mitigating Catastrophic Forgetting in Language Transfer via Model Merging.
\newblock In Al-Onaizan, Y.; Bansal, M.; and Chen, Y.-N., eds., \emph{Findings of the Association for Computational Linguistics: EMNLP 2024}, 17167--17186. Miami, Florida, USA: Association for Computational Linguistics.

\bibitem[{Chen and Lee(2024)}]{chen2024instructioncp}
Chen, K.-M.; and Lee, H.-y. 2024.
\newblock InstructionCP: A fast approach to transfer Large Language Models into target language.
\newblock \emph{arXiv preprint arXiv:2405.20175}.

\bibitem[{Clark et~al.(2019)Clark, Lee, Chang, Kwiatkowski, Collins, and Toutanova}]{clark-etal-2019-boolq}
Clark, C.; Lee, K.; Chang, M.-W.; Kwiatkowski, T.; Collins, M.; and Toutanova, K. 2019.
\newblock {B}ool{Q}: Exploring the Surprising Difficulty of Natural Yes/No Questions.
\newblock In Burstein, J.; Doran, C.; and Solorio, T., eds., \emph{Proceedings of the 2019 Conference of the North {A}merican Chapter of the Association for Computational Linguistics: Human Language Technologies, Volume 1 (Long and Short Papers)}, 2924--2936. Minneapolis, Minnesota: Association for Computational Linguistics.

\bibitem[{Dou et~al.(2024)Dou, Liu, Zeng, Guo, Zhou, Mao, Jin, Lu, and Lin}]{dou-etal-2024-sailor}
Dou, L.; Liu, Q.; Zeng, G.; Guo, J.; Zhou, J.; Mao, X.; Jin, Z.; Lu, W.; and Lin, M. 2024.
\newblock Sailor: Open Language Models for South-{E}ast {A}sia.
\newblock In Hernandez~Farias, D.~I.; Hope, T.; and Li, M., eds., \emph{Proceedings of the 2024 Conference on Empirical Methods in Natural Language Processing: System Demonstrations}, 424--435. Miami, Florida, USA: Association for Computational Linguistics.

\bibitem[{Fang et~al.(2021)Fang, Wang, Wang, Zhang, Yang, and Liu}]{fang2021seed}
Fang, Z.; Wang, J.; Wang, L.; Zhang, L.; Yang, Y.; and Liu, Z. 2021.
\newblock {\{}SEED{\}}: Self-supervised Distillation For Visual Representation.
\newblock In \emph{International Conference on Learning Representations}.

\bibitem[{Fujii et~al.(2024)Fujii, Nakamura, Loem, Iida, Ohi, Hattori, Shota, Mizuki, Yokota, and Okazaki}]{fujiicontinual}
Fujii, K.; Nakamura, T.; Loem, M.; Iida, H.; Ohi, M.; Hattori, K.; Shota, H.; Mizuki, S.; Yokota, R.; and Okazaki, N. 2024.
\newblock Continual Pre-Training for Cross-Lingual LLM Adaptation: Enhancing Japanese Language Capabilities.
\newblock In \emph{First Conference on Language Modeling}.

\bibitem[{Fujinuma, Boyd-Graber, and Kann(2022)}]{fujinuma-etal-2022-match}
Fujinuma, Y.; Boyd-Graber, J.; and Kann, K. 2022.
\newblock Match the Script, Adapt if Multilingual: Analyzing the Effect of Multilingual Pretraining on Cross-lingual Transferability.
\newblock In Muresan, S.; Nakov, P.; and Villavicencio, A., eds., \emph{Proceedings of the 60th Annual Meeting of the Association for Computational Linguistics (Volume 1: Long Papers)}, 1500--1512. Dublin, Ireland: Association for Computational Linguistics.

\bibitem[{Guo et~al.(2024)Guo, Ren, Hu, Li, Zhang, Zhang, and Huang}]{guo-etal-2024-teaching}
Guo, P.; Ren, Y.; Hu, Y.; Li, Y.; Zhang, J.; Zhang, X.; and Huang, H. 2024.
\newblock Teaching Large Language Models to Translate on Low-resource Languages with Textbook Prompting.
\newblock In Calzolari, N.; Kan, M.-Y.; Hoste, V.; Lenci, A.; Sakti, S.; and Xue, N., eds., \emph{Proceedings of the 2024 Joint International Conference on Computational Linguistics, Language Resources and Evaluation (LREC-COLING 2024)}, 15685--15697. Torino, Italia: ELRA and ICCL.

\bibitem[{Hinton, Vinyals, and Dean(2015)}]{hinton2015distilling}
Hinton, G.; Vinyals, O.; and Dean, J. 2015.
\newblock Distilling the knowledge in a neural network.
\newblock \emph{arXiv preprint arXiv:1503.02531}.

\bibitem[{Jang et~al.(2022)Jang, Kim, Kwon, and Davis}]{jang2022kobest}
Jang, M.; Kim, D.; Kwon, D.~S.; and Davis, E. 2022.
\newblock {K}o{BEST}: {K}orean Balanced Evaluation of Significant Tasks.
\newblock In Calzolari, N.; Huang, C.-R.; Kim, H.; Pustejovsky, J.; Wanner, L.; Choi, K.-S.; Ryu, P.-M.; Chen, H.-H.; Donatelli, L.; Ji, H.; Kurohashi, S.; Paggio, P.; Xue, N.; Kim, S.; Hahm, Y.; He, Z.; Lee, T.~K.; Santus, E.; Bond, F.; and Na, S.-H., eds., \emph{Proceedings of the 29th International Conference on Computational Linguistics}, 3697--3708. Gyeongju, Republic of Korea: International Committee on Computational Linguistics.

\bibitem[{Kim et~al.(2021)Kim, Kim, Lee, Lee, Kwak, Dong~Hyeon, Park, Kim, Kim, Seo, Lee, Jeong, Lee, Kim, Ko, Kim, Park, Kim, Kang, Ryu, Yoo, Chang, Suh, In, Park, Kim, Kim, Jeong, Yeo, Ham, Park, Lee, Kang, Kang, Ha, Park, and Sung}]{kim-etal-2021-changes}
Kim, B.; Kim, H.; Lee, S.-W.; Lee, G.; Kwak, D.; Dong~Hyeon, J.; Park, S.; Kim, S.; Kim, S.; Seo, D.; Lee, H.; Jeong, M.; Lee, S.; Kim, M.; Ko, S.~H.; Kim, S.; Park, T.; Kim, J.; Kang, S.; Ryu, N.-H.; Yoo, K.~M.; Chang, M.; Suh, S.; In, S.; Park, J.; Kim, K.; Kim, H.; Jeong, J.; Yeo, Y.~G.; Ham, D.; Park, D.; Lee, M.~Y.; Kang, J.; Kang, I.; Ha, J.-W.; Park, W.; and Sung, N. 2021.
\newblock What Changes Can Large-scale Language Models Bring? Intensive Study on {H}yper{CLOVA}: Billions-scale {K}orean Generative Pretrained Transformers.
\newblock In Moens, M.-F.; Huang, X.; Specia, L.; and Yih, S. W.-t., eds., \emph{Proceedings of the 2021 Conference on Empirical Methods in Natural Language Processing}, 3405--3424. Online and Punta Cana, Dominican Republic: Association for Computational Linguistics.

\bibitem[{Kim, Hirasawa, and Komachi(2020)}]{kim-etal-2020-korean}
Kim, H.; Hirasawa, T.; and Komachi, M. 2020.
\newblock {K}orean-to-{J}apanese Neural Machine Translation System using Hanja Information.
\newblock In Nakazawa, T.; Nakayama, H.; Ding, C.; Dabre, R.; Kunchukuttan, A.; Pa, W.~P.; Bojar, O.; Parida, S.; Goto, I.; Mino, H.; Manabe, H.; Sudoh, K.; Kurohashi, S.; and Bhattacharyya, P., eds., \emph{Proceedings of the 7th Workshop on Asian Translation}, 127--134. Suzhou, China: Association for Computational Linguistics.

\bibitem[{Kim et~al.(2024)Kim, Park, Kim, and Lee}]{kim-etal-2024-kombo}
Kim, S.; Park, J.; Kim, Y.; and Lee, S. 2024.
\newblock {KOMBO}: {K}orean Character Representations Based on the Combination Rules of Subcharacters.
\newblock In Ku, L.-W.; Martins, A.; and Srikumar, V., eds., \emph{Findings of the Association for Computational Linguistics: ACL 2024}, 5102--5119. Bangkok, Thailand: Association for Computational Linguistics.

\bibitem[{Lee et~al.(2024)Lee, Moon, Lee, Park, Eo, Ko, Seo, Lee, and Lim}]{lee-etal-2024-length}
Lee, J.; Moon, H.; Lee, S.; Park, C.; Eo, S.; Ko, H.; Seo, J.; Lee, S.; and Lim, H. 2024.
\newblock Length-aware Byte Pair Encoding for Mitigating Over-segmentation in {K}orean Machine Translation.
\newblock In Ku, L.-W.; Martins, A.; and Srikumar, V., eds., \emph{Findings of the Association for Computational Linguistics: ACL 2024}, 2287--2303. Bangkok, Thailand: Association for Computational Linguistics.

\bibitem[{Li, Sha, and Shi(2020)}]{li2020revisiting}
Li, H.; Sha, J.; and Shi, C. 2020.
\newblock Revisiting back-translation for low-resource machine translation between Chinese and Vietnamese.
\newblock \emph{IEEE Access}, 8: 119931--119939.

\bibitem[{Liu, Winata, and Fung(2021)}]{liu-etal-2021-continual}
Liu, Z.; Winata, G.~I.; and Fung, P. 2021.
\newblock Continual Mixed-Language Pre-Training for Extremely Low-Resource Neural Machine Translation.
\newblock In Zong, C.; Xia, F.; Li, W.; and Navigli, R., eds., \emph{Findings of the Association for Computational Linguistics: ACL-IJCNLP 2021}, 2706--2718. Online: Association for Computational Linguistics.

\bibitem[{Luo et~al.(2024)Luo, Zhang, Liu, Li, Gong, Qi, and Cheng}]{luo2024velocitune}
Luo, Z.; Zhang, X.; Liu, X.; Li, H.; Gong, Y.; Qi, C.; and Cheng, P. 2024.
\newblock Velocitune: A Velocity-based Dynamic Domain Reweighting Method for Continual Pre-training.
\newblock \emph{arXiv preprint arXiv:2411.14318}.

\bibitem[{Moosa, Akhter, and Habib()}]{moosatransliteration}
Moosa, I.~M.; Akhter, M.~E.; and Habib, A.~B. ????
\newblock Transliteration: A Simple Technique For Improving Multilingual Language Modeling.

\bibitem[{Nag et~al.(2025)Nag, Chakrabarti, Mukherjee, and Ganguly}]{nag-etal-2025-efficient}
Nag, A.; Chakrabarti, S.; Mukherjee, A.; and Ganguly, N. 2025.
\newblock Efficient Continual Pre-training of {LLM}s for Low-resource Languages.
\newblock In Chen, W.; Yang, Y.; Kachuee, M.; and Fu, X.-Y., eds., \emph{Proceedings of the 2025 Conference of the Nations of the Americas Chapter of the Association for Computational Linguistics: Human Language Technologies (Volume 3: Industry Track)}, 304--317. Albuquerque, New Mexico: Association for Computational Linguistics.
\newblock ISBN 979-8-89176-194-0.

\bibitem[{Navigli(2009)}]{10.1145/1459352.1459355}
Navigli, R. 2009.
\newblock Word sense disambiguation: A survey.
\newblock \emph{ACM Comput. Surv.}, 41(2).

\bibitem[{Ogueji, Zhu, and Lin(2021)}]{ogueji-etal-2021-small}
Ogueji, K.; Zhu, Y.; and Lin, J. 2021.
\newblock Small Data? No Problem! Exploring the Viability of Pretrained Multilingual Language Models for Low-resourced Languages.
\newblock In Ataman, D.; Birch, A.; Conneau, A.; Firat, O.; Ruder, S.; and Sahin, G.~G., eds., \emph{Proceedings of the 1st Workshop on Multilingual Representation Learning}, 116--126. Punta Cana, Dominican Republic: Association for Computational Linguistics.

\bibitem[{Park and Zhao(2019)}]{park2019korean}
Park, J.; and Zhao, H. 2019.
\newblock Korean-to-chinese machine translation using chinese character as pivot clue.
\newblock \emph{arXiv preprint arXiv:1911.11008}.

\bibitem[{Penedo et~al.(2024)Penedo, Kydl{\'\i}{\v{c}}ek, Lozhkov, Mitchell, Raffel, Von~Werra, Wolf et~al.}]{penedo2024fineweb}
Penedo, G.; Kydl{\'\i}{\v{c}}ek, H.; Lozhkov, A.; Mitchell, M.; Raffel, C.~A.; Von~Werra, L.; Wolf, T.; et~al. 2024.
\newblock The fineweb datasets: Decanting the web for the finest text data at scale.
\newblock \emph{Advances in Neural Information Processing Systems}, 37: 30811--30849.

\bibitem[{Pilehvar and Camacho-Collados(2019)}]{pilehvar-camacho-collados-2019-wic}
Pilehvar, M.~T.; and Camacho-Collados, J. 2019.
\newblock {W}i{C}: the Word-in-Context Dataset for Evaluating Context-Sensitive Meaning Representations.
\newblock In Burstein, J.; Doran, C.; and Solorio, T., eds., \emph{Proceedings of the 2019 Conference of the North {A}merican Chapter of the Association for Computational Linguistics: Human Language Technologies, Volume 1 (Long and Short Papers)}, 1267--1273. Minneapolis, Minnesota: Association for Computational Linguistics.

\bibitem[{Raju et~al.(2025)Raju, Walia, Raghav, Marivate et~al.}]{raju2025afroxlmr}
Raju, J.~S.; Walia, J.~S.; Raghav, S.; Marivate, V.; et~al. 2025.
\newblock AfroXLMR-Comet: Multilingual Knowledge Distillation with Attention Matching for Low-Resource languages.
\newblock \emph{arXiv preprint arXiv:2502.18020}.

\bibitem[{Roemmele, Bejan, and Gordon(2011)}]{roemmele2011choice}
Roemmele, M.; Bejan, C.~A.; and Gordon, A.~S. 2011.
\newblock Choice of Plausible Alternatives: An Evaluation of Commonsense Causal Reasoning.
\newblock In \emph{AAAI spring symposium: logical formalizations of commonsense reasoning}, 90--95.

\bibitem[{Shin et~al.(2025)Shin, Lee, Jang, Song, Kim, Oh, Jo, Ahn, Oh, Chang et~al.}]{shin2025kobalt}
Shin, H.; Lee, S.; Jang, D.; Song, W.; Kim, J.; Oh, C.; Jo, H.; Ahn, Y.; Oh, S.; Chang, H.; et~al. 2025.
\newblock KoBALT: Korean Benchmark For Advanced Linguistic Tasks.
\newblock \emph{arXiv preprint arXiv:2505.16125}.

\bibitem[{Tao et~al.(2024)Tao, Zhang, Huang, Ma, Huang, Zhao, and Feng}]{tao-etal-2024-unlocking}
Tao, M.; Zhang, C.; Huang, Q.; Ma, T.; Huang, S.; Zhao, D.; and Feng, Y. 2024.
\newblock Unlocking the Potential of Model Merging for Low-Resource Languages.
\newblock In Al-Onaizan, Y.; Bansal, M.; and Chen, Y.-N., eds., \emph{Findings of the Association for Computational Linguistics: EMNLP 2024}, 8705--8720. Miami, Florida, USA: Association for Computational Linguistics.

\bibitem[{Vo et~al.(2024)Vo, Jung, Lee, and Choi}]{vo2024redwhale}
Vo, A.-D.; Jung, M.; Lee, W.; and Choi, D. 2024.
\newblock Redwhale: An adapted korean llm through efficient continual pretraining.
\newblock \emph{arXiv preprint arXiv:2408.11294}.

\bibitem[{Xu et~al.(2024)Xu, Li, Tao, Shen, Cheng, Li, Xu, Tao, and Zhou}]{xu2024survey}
Xu, X.; Li, M.; Tao, C.; Shen, T.; Cheng, R.; Li, J.; Xu, C.; Tao, D.; and Zhou, T. 2024.
\newblock A survey on knowledge distillation of large language models.
\newblock \emph{arXiv preprint arXiv:2402.13116}.

\bibitem[{Yoo, Kim, and Lee(2019)}]{yoo-etal-2019-dont}
Yoo, K.~M.; Kim, T.; and Lee, S.-g. 2019.
\newblock Don{'}t Just Scratch the Surface: Enhancing Word Representations for {K}orean with Hanja.
\newblock In Inui, K.; Jiang, J.; Ng, V.; and Wan, X., eds., \emph{Proceedings of the 2019 Conference on Empirical Methods in Natural Language Processing and the 9th International Joint Conference on Natural Language Processing (EMNLP-IJCNLP)}, 3528--3533. Hong Kong, China: Association for Computational Linguistics.

\bibitem[{Zellers et~al.(2019)Zellers, Holtzman, Bisk, Farhadi, and Choi}]{zellers-etal-2019-hellaswag}
Zellers, R.; Holtzman, A.; Bisk, Y.; Farhadi, A.; and Choi, Y. 2019.
\newblock {H}ella{S}wag: Can a Machine Really Finish Your Sentence?
\newblock In Korhonen, A.; Traum, D.; and M{\`a}rquez, L., eds., \emph{Proceedings of the 57th Annual Meeting of the Association for Computational Linguistics}, 4791--4800. Florence, Italy: Association for Computational Linguistics.

\end{thebibliography}

\end{document}